\documentclass{article}

\PassOptionsToPackage{table}{xcolor}
\usepackage[utf8]{inputenc}

\usepackage{graphicx}

\usepackage[parfill]{parskip}
\usepackage{amsmath,amsthm}
\usepackage{mathtools}
\usepackage[citestyle=authoryear-comp, uniquename=false, uniquelist=false,sorting=ynt, maxbibnames=99, maxcitenames=1, backend=biber]{biblatex}
\newcommand{\citep}{\parencite}

\newcommand{\citet}{\textcite}
\RequirePackage[tt=false, type1=true]{libertine}
\RequirePackage[varqu]{zi4}
\RequirePackage[libertine]{newtxmath}

\usepackage{bm}
\usepackage[inline]{enumitem}
\usepackage{booktabs}
\usepackage{algorithm,algorithmicx,algpseudocode}
\usepackage{nicematrix}

\usepackage[T1]{fontenc}
\usepackage{url}
\usepackage{amsfonts}
\usepackage{microtype}

\usepackage[colorlinks]{hyperref}
\usepackage[capitalise,noabbrev]{cleveref}

\definecolor{c7}{HTML}{6C7869}
\definecolor{c6}{HTML}{FBE7B4}
\definecolor{c5}{HTML}{9B5353}
\definecolor{c1}{HTML}{586770}
\definecolor{c4}{HTML}{2a4a67}
\definecolor{c3}{HTML}{6d2a58}
\definecolor{c2}{HTML}{34142a}

\definecolor{mydarkred}{HTML}{9B5353}
\definecolor{mydarkpurple}{HTML}{65657E}
\definecolor{mydarkgreen}{HTML}{6C7869}
\definecolor{darkred}{HTML}{a70c0c}
\definecolor{darkblue}{HTML}{4D5B9C}
\definecolor{myblue}{HTML}{FDF5E0} 
\definecolor{mygray}{HTML}{DBE2E9} 
\definecolor{mygreen}{HTML}{E6F3FC} 
\definecolor{dark2orange}{rgb}{0.9, 0.4, 0.}
\definecolor{dark2purple}{rgb}{0.4, 0.4, 0.8}

\hypersetup{
    colorlinks=true,
    linkcolor=black,
    urlcolor=c1,
    citecolor=c1,
}

\newcommand{\M}[0]{\mathcal{M}}
\newcommand{\G}[0]{\mathcal{G}}
\newcommand{\head}[1]{\vspace{1.7mm}\noindent{{\textcolor{c4}{\bf #1.}}}}
\newcommand{\model}[0]{\textsc{Proteus}}
\newtheorem{dfn}{Definition}
\crefname{dfn}{Definition}{Definitions}
\Crefname{dfn}{Definition}{Definitions}

\usepackage{authblk}

\newcommand{\vk}{\boldsymbol{k}}
\newcommand{\vq}{\boldsymbol{q}}
\renewcommand{\vv}{\boldsymbol{v}}

\newcommand{\vx}{\boldsymbol{x}}
\newcommand{\vy}{\boldsymbol{y}}

\newcommand\blfootnote[1]{%
  \begingroup
  \footnotetext{#1}%
  \endgroup
}

\title{\model{}: Incremental Memory Activation for Long-Context \\ Sequence Modeling} 

\author{Reza Bayat$^{\dagger *}$ \protect \blfootnote{Correspondence
to: \texttt{\{reza.bayat,~courvila\}@mila.quebec} \: and \: \texttt{\{alibehrouz,~mirrokni\}@google.com}.}\!\!\protect \blfootnote{$^*$ Equal contribution.}}
\author{Ali Behrouz$^{\ddagger *}$}
\author{Vahab Mirrokni$^{\ddagger}$}
\author{Aaron Courville$^{\dagger}$}
 
\affil[]{%
  \centering
  \makebox[6cm][c]{\protect \Large $^{^{^\dagger}}$\includegraphics[height=6mm]{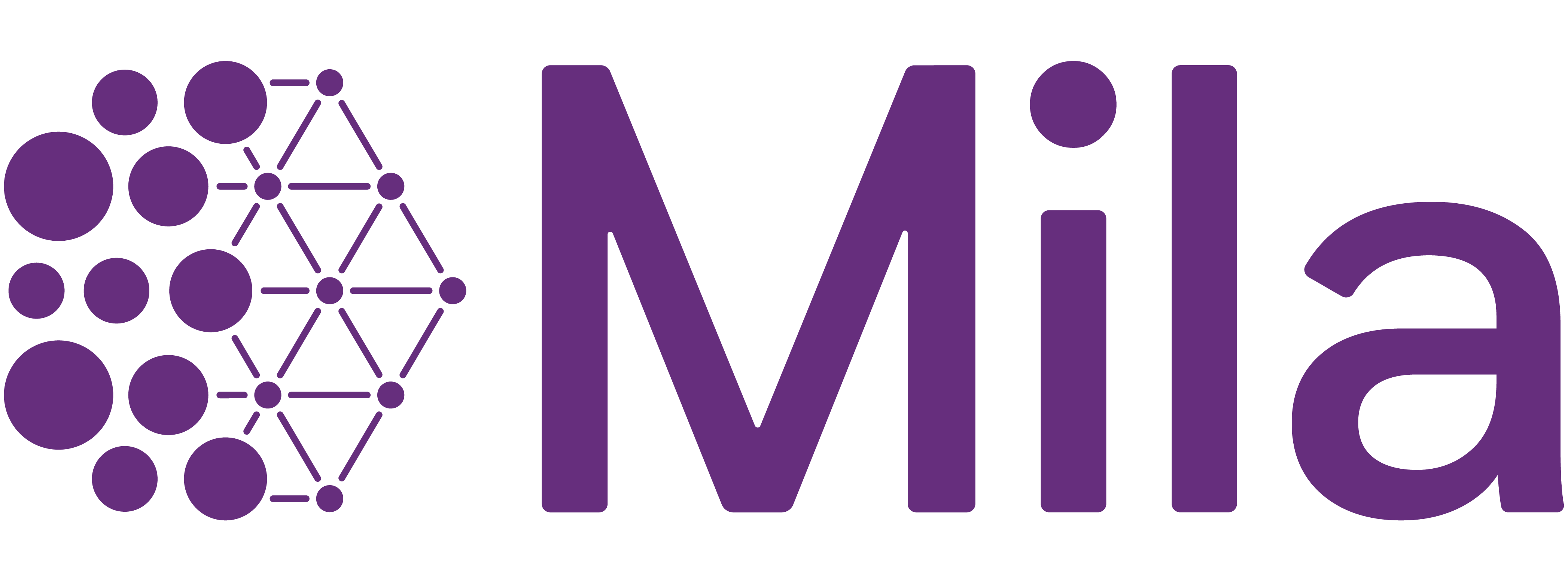}}%
  \makebox[5.75cm][c]{\protect \Large $^{^{^\ddagger}}$\includegraphics[height=5.75mm]{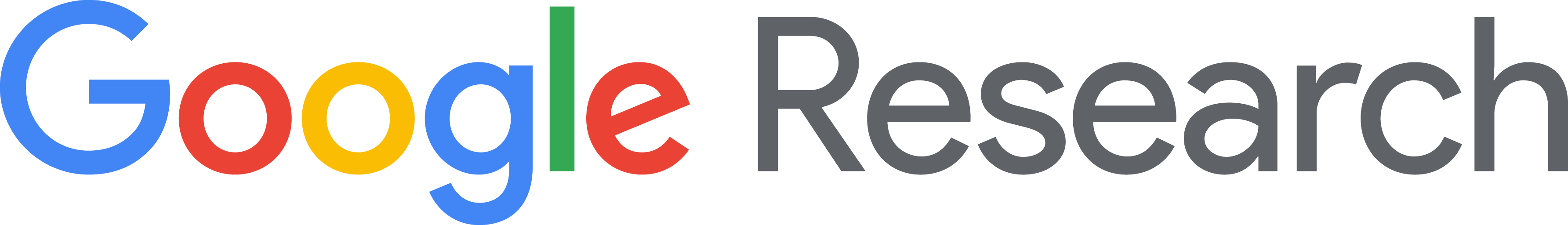}}%
}
\date{}

\begin{document}

\maketitle

\begin{abstract}
The quadratic cost of attention-based sequence models for long contexts has motivated a growing line of research on \emph{memory-based} models that can compress context into a compact state.
However, most existing memory models expose a \emph{static} memory throughout the entire sequence.
Because early tokens face no compression pressure, they occupy too many degrees of freedom and ``pollute'' the memory state, leaving little capacity for later context and increasing interference between what is stored and what arrives next.
We study a new paradigm of \emph{incremental memory activation}, where the effective capacity of memory is progressively expanded as the context grows.
Imposing an early bottleneck forces the model to compress history more effectively, while unlocking fresh capacity over time reduces interference and improves retention of later context.
We instantiate this paradigm in \model{}, a straightforward mechanism that can be incorporated into a broad class of neural memory architectures at no additional cost.
We apply \model{} to state-of-the-art models, including SWLA, Comba, Titans, and Hope-Attention, and observe consistent improvements on standard language modeling and reasoning, as well as on long-context retrieval and understanding, with gains that grow at longer context lengths.
Overall, our results show that static memory is suboptimal and that scheduling effective capacity is a simple and broadly applicable tool for sequence modeling.
\end{abstract}

\vspace{4ex}
\begin{figure}[H]
    \centering
    \includegraphics[width=1.0\linewidth]{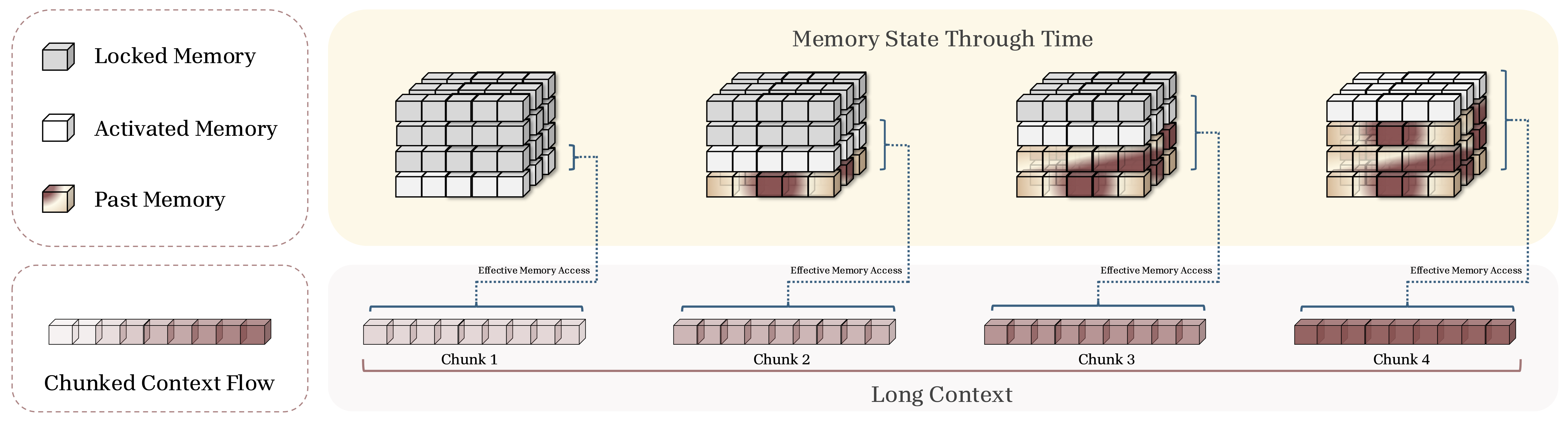}
    \caption{\textbf{Incremental memory activation in \model{}.} \model{} progressively expands the active subset of the memory state as the context advances. At each step, the model both reads from and writes to the currently \emph{activated} blocks (together with previously activated ``past'' blocks), while the remaining blocks stay \emph{locked} and do not participate in retrieval or updates. As the context grows, additional blocks are unlocked, providing fresh capacity for later tokens. This imposes an early capacity bottleneck that encourages compression of the initial context, while the fresh capacity introduced over time reduces overwriting and interference for later tokens.}
    \label{fig:proteus-arc}
\end{figure}

\newpage
\section{Introduction}
The rise of Transformers~\citep{Vaswani+2017}, pure attention-based architectures, has revolutionized the use cases of machine learning and AI, moving from task-specific models to general-purpose systems~\citep{brown2020language, kaplan2020scaling}. Softmax attention, the backbone of Transformers and the component mainly responsible for sequence mixing, acts as a lossless memory module that caches every single token in the sequence, so that its memory size grows linearly with sequence length~\citep{ramsauer2021hopfield, bietti2024birth, behrouz2025Miras}. While this makes attention well suited to retrieval, the growing memory comes with a quadratic computational cost with respect to the sequence length, limiting the model's long-context handling.

To overcome the quadratic cost of Transformers, as well as their limited expressivity on tasks that require sequential reasoning, modern recurrent neural networks have re-gained popularity in recent years~\citep{katharopoulos2020transformers, irie2021going, sun2023retentive, sun2024learning, behrouz2024titans}. In contrast to Transformers with their growing memory, modern RNNs employ a fixed-size memory (a.k.a. a hidden state) to compress the context, so that the decoding cost per token is constant and the overall computational cost is linear or sub-quadratic in the sequence length. This efficiency, however, comes at a price: compressing an unbounded context into a fixed-size state limits the length extrapolation ability of RNNs---their ability to generalize to sequence lengths beyond those seen in training---and thus their potential for long-context understanding~\citep{bai2023longbench0, kuratov2024babilong, hsieh2024ruler, tiezzi2024resurgence}.

A substantial body of work has sought to close this gap through (meta-learning) memory initialization~\citep{sun2024learning, behrouz2024titans}, more robust update rules~\citep{schlag2021linear, von2023uncovering, behrouz2025atlas}, and more powerful memory architectures~\citep{sun2024learning, behrouz2024titans, zhang2025test}. Despite these advances, recurrent models still struggle to make full use of their memory capacity, and we argue that a key reason is \emph{when}, not just how, information is written. Because the memory is written online, early tokens arrive when the state is nearly empty: they face little competition for capacity and are under almost no pressure to compress, and so they are free to occupy a disproportionate share of the memory's degrees of freedom. Later tokens, by contrast, must be squeezed into whatever capacity remains, and writing them increasingly overwrites or interferes with what is already stored. The result is an imbalance in which the memory is biased toward initial tokens~\citep{barbero2024transformers, behrouz2025best} and struggles to incorporate later context---precisely the regime that long-context tasks stress most.

In this paper, we study a paradigm of \emph{incremental memory activation} for recurrent neural networks, in which the effective memory capacity is scheduled as a function of context length. The design directly targets the imbalance above. Since the problem originates with early tokens being under-compressed, we deliberately impose a capacity bottleneck early in the sequence, forcing the model to summarize the initial context rather than memorize it; and since later tokens suffer from interference, we progressively unlock fresh, previously unused capacity as the context grows, giving new information room to be written without overwriting what came before. In this way, the two mechanisms---an early bottleneck and fresh capacity over time---address the two failure modes of static memory in tandem. More broadly, viewing MLP-block through the Nested Learning lens~\citep{behrouz2025nested} as a form of associative memory, the same scheduling principle extends beyond the recurrent state to a model's parameters, which lets us apply it to architectures such as Hope-Attention.

We instantiate this paradigm with a simple yet effective mechanism, \model{}, that can be integrated into modern deep and linear-memory recurrent models at no additional cost. Across a diverse set of benchmarks spanning language modeling, commonsense reasoning, long-context understanding, and needle-in-a-haystack tasks, \model{} improves over strong baselines on both standard and long-context settings, with gains that grow at longer context lengths.

\head{Contributions} In summary, our key contributions in this paper are as follows.
\begin{itemize}[leftmargin=*, itemsep=2pt]
    \item \textbf{Incremental memory activation.} We introduce incremental memory activation, a paradigm for controlling \emph{effective} model capacity (e.g., memory or parameters) over the context flow via progressive activation, encouraging early compression while reducing later overwrite and interference.
    \item \textbf{\model{}: a general, lightweight mechanism.} We propose \model{}, a simple block-wise gating scheme that applies incremental activation to both memory reads and writes and is drop-in for a broad class of associative-memory architectures.
    \item \textbf{Strong results.} We show that \model{} consistently improves state-of-the-art models (SWLA, Comba, Titans, Hope-Attention) across model scales on a broad range of benchmarks, without adding parameters or memory.
\end{itemize}

\section{Preliminaries}
\label{sec:preliminaries}

\subsection{Sequence Modeling as Associative Memory}
A growing line of work views sequence models through the lens of an \emph{associative memory system} that compresses past context into a compact state via an optimization procedure~\citep{sun2024learning, behrouz2024titans, behrouz2025Miras}. We adopt the general formulation of~\citet{behrouz2025Miras}.

\begin{dfn}[Associative Memory]\label{def:associative-memory}
Given a set of keys $\mathcal{K} \subseteq \mathbb{R}^{d_k}$ and values $\mathcal{V} \subseteq \mathbb{R}^{d_v}$, an associative memory is an operator $\M(\cdot)$, parameterized by a set of memory parameters, that maps the keys $\mathcal{K}$ to values in $\mathcal{V}$. To learn such a mapping from data, an internal objective $\tilde{\mathcal{L}}(\cdot;\cdot)$ measures the quality of the mapping, and $\M$ can be computed by solving:
\begin{align}\label{eq:attentional-bias-loss}
    \M^* = \arg\min_{\M}\; \tilde{\mathcal{L}}\bigl(\M(\mathcal{K}), \mathcal{V}\bigr).
\end{align}
\end{dfn}

In the sequence modeling setting, $\mathcal{K}$ and $\mathcal{V}$ are constructed from different representation views of the inputs, typically linear projections $k_t = x_t W_k$, $v_t = x_t W_v$, and $q_t = x_t W_q$~\citep{sun2024learning, behrouz2024titans}. We use the tilde in $\tilde{\mathcal{L}}$ to distinguish this \emph{internal} memory objective---which the memory minimizes to learn its key--value associations---from the outer training loss of the full model~\citep{behrouz2025Miras}. The power of this view is that a single choice of internal objective and optimization procedure induces a specific architecture: different sequence models are not fundamentally different mechanisms, but different instantiations of \cref{eq:attentional-bias-loss}~\citep{behrouz2025nested}.

\head{A spectrum of compression}
This framing situates existing sequence models along a spectrum defined by how aggressively they compress context. At one extreme, softmax attention can be cast as the \emph{non-parametric} solution to an $\ell_2$ regression objective over the context, storing every key--value pair without compression and thus acting as a lossless but linearly growing memory; restricting that objective to a local window recovers sliding-window attention~\citep{behrouz2025nested}. At the other extreme, a fixed-size recurrent memory compresses the entire history into a state of constant size, trading expressivity for efficiency. Incremental memory activation, introduced in \cref{sec:incremental_memory_activation}, can be seen as modulating \emph{where on this spectrum} the effective memory sits as the context grows.

\subsection{Memory as Online Optimization}
\label{sec:online-memory}
The memory in \cref{def:associative-memory} is learned \emph{online}, one token at a time, as the context is revealed. Concretely, the recurrence of a memory-based model can be written as an update rule that writes each new input into the state, and a read rule that produces a memory-conditioned output. A common instantiation optimizes the internal objective by gradient descent, giving the update
\begin{align}\label{eq:generic-update}
    \M_t = \M_{t-1} - \theta_t\, \nabla\tilde{\mathcal{L}}(\M_{t-1}; k_t, v_t),
\end{align}
where $\theta_t$ is a step size, together with a read $y_t = \mathrm{Read}(\M_{t-1}, q_t)$ that depends on the memory architecture (e.g., a matrix--vector product for a linear memory, or a forward pass for an MLP memory). The gradient $\nabla\tilde{\mathcal{L}}(\M_{t-1}; k_t, v_t)$ acts as a \emph{surprise} signal: the more the current association departs from what the memory has captured, the larger the update~\citep{behrouz2024titans}.

\head{Architectures as choices of objective and update}
Within this frame, distinct architectures correspond to distinct choices of the internal objective $\tilde{\mathcal{L}}$, the optimizer (including retention and momentum), and the memory parameterization~\citep{behrouz2025Miras, behrouz2025nested}. 

\emph{Hebbian rule.} Taking a dot-product objective $\tilde{\mathcal{L}}(\M_{t-1}; k_t, v_t) = -\langle \M_{t-1} k_t, v_t\rangle$ and optimizing by gradient descent with a decay term recovers linear attention~\citep{katharopoulos2020transformers}:
\begin{align}\label{eq:hebbian}
    \M_t = \alpha_t\, \M_{t-1} + \theta_t\, v_t k_t^{\top}.
\end{align}
The additive write has limited capacity, since it never removes stale associations before writing new ones.

\emph{Delta rule.} Replacing the objective with an $\ell_2$ reconstruction loss $\tilde{\mathcal{L}}(\M_{t-1}; k_t, v_t) = \tfrac{1}{2}\|\M_{t-1} k_t - v_t\|_2^2$ yields a delta-rule update that first erases the current value along $k_t$ before writing $v_t$~\citep{schlag2021linear}:
\begin{align}\label{eq:delta}
    \M_t = \bigl(I - \theta_t\, k_t k_t^{\top}\bigr)\M_{t-1} + \theta_t\, v_t k_t^{\top}.
\end{align}

\emph{Momentum and forgetting (Titans).} Building on the $\ell_2$ objective, Titans augments the update with momentum and an adaptive forget gate~\citep{behrouz2024titans}:
\begin{align}\label{eq:titans}
    \M_t = (1 - \alpha_t)\,\M_{t-1} + S_t, \qquad S_t = \eta_t\, S_{t-1} - \theta_t\, \nabla\tilde{\mathcal{L}}(\M_{t-1}; k_t, v_t), 
\end{align}
where $S_t$ carries a running memory of past surprise (momentum) and $\alpha_t \in [0,1]$ controls how much of the state is forgotten. TTT~\citep{sun2024learning} corresponds to the same $\ell_2$ objective with plain gradient descent ($\eta_t = \alpha_t = 0$) and admits a linear or an MLP memory.

\head{Effective capacity}
We refer to the \emph{effective capacity} of the memory at step $t$ as the number of memory parameters that are active---i.e., that participate in the update in \cref{eq:generic-update} and in the read. We write $C$ for the total capacity, and in \cref{sec:incremental_memory_activation} we schedule an active subset $c(t) \le C$ over the context. This axis is deliberately \emph{orthogonal} to the dimensions discussed in general frameworks~\citep{behrouz2025Miras, behrouz2025nested, wang2025test}: incremental activation changes neither the internal objective, nor the optimizer, nor the memory architecture, but only which parameters are exposed at each position. This orthogonality is precisely what lets a single mechanism apply, unchanged, across the entire family.

\subsection{MLP Blocks as Associative Memory}
\label{sec:nl-pretraining}
The associative-memory view above concerns the online memory state, but a recent perspective---Nested Learning~\citep{behrouz2025nested}---extends the same lens to a model's \emph{parameters}. Under this view, gradient-based update of MLP blocks is itself a form of associative memory: each layer, updated by backpropagation, learns to map its input to the local error signal it receives. Concretely, a gradient-descent update
\begin{align}\label{eq:nl-backprop}
    \bm{\theta}_{i+1} = \bm{\theta}_i - \eta_{i+1}\, \bm{e}_i,
\end{align}
where $\bm{e}_i$ is the error (surprise) supplied by the optimizer for the $i$-th data sample, can be read as a memory that writes each sample's contribution into $\bm{\theta}$. We do not rely on the full machinery of this perspective; what matters for us is the shared structure. This motivates applying the same incremental-activation principle not only to the recurrent state but to the MLP parameters themselves, which we develop for architectures such as Hope-Attention in \cref{sec:proteus}.

\subsection{Static Capacity is Suboptimal}
Despite their variety, the memory and parameter views above share a design choice that is rarely questioned: capacity is \emph{static}, with the same parameters exposed throughout the entire context or training run. Yet controlling a model's effective capacity has long been studied~\citep{guyon1991structural, hansen2001model, ding2018model, kaplan2020scaling}, and from a compression standpoint, a bottlenecked information flow encourages a model to retain only task-relevant structure, as exemplified by autoencoders~\citep{vincent2010stacked}. Classical capacity control restricts a model's degrees of freedom \emph{globally}; the online setting of \cref{eq:generic-update} suggests a positional analogue, since the right amount of capacity differs early versus late.

The failure mode of static capacity can be read directly off the online objective. When all parameters of $\M$ are active from the first token, early inputs are fit using the memory's full degrees of freedom; minimizing $\tilde{\mathcal{L}}$ then favors memorizing them rather than compressing them, so the memory is ``polluted'' by early context and generalizes poorly. Once the state is heavily written, updating it for later tokens increasingly interferes with what is already stored, making new information harder to incorporate. In the following, we introduce \emph{incremental memory activation} for controlling effective capacity over the context in \cref{sec:incremental_memory_activation}, instantiate it in \model{} in \cref{sec:proteus}, and incorporate it into a broad class of sequence models in \cref{sec:experiments}.

\section{Incremental Memory Activation}
\label{sec:incremental_memory_activation}
\cref{sec:preliminaries} identified two failure modes of static capacity: early inputs are memorized rather than compressed, and later inputs interfere with an already-saturated state. We now introduce a paradigm that addresses both by scheduling the \emph{effective} capacity of the memory as a function of input position. The design follows two complementary goals: (i) \textbf{early compression}, which forces the model to summarize early context through a restricted-capacity bottleneck; and (ii) \textbf{fresh capacity}, which unlocks previously unused components later in the context to absorb new information without overwriting what has already been stored.

\head{Why a growing schedule}
Before formalizing the paradigm, it is worth asking what kind of schedule the two goals call for. The tension is between bias and interference. Too little active capacity late in the context wastes tokens the memory could have used, while too much active capacity early lets the memory overfit the few tokens seen so far instead of compressing them. Balancing the two suggests that the active capacity should track the amount of context observed: small when few tokens have been seen, and growing as more arrive, rather than being fixed at either extreme. This intuition, that effective capacity should grow with position, motivates the simple uniform schedule we adopt in \cref{sec:proteus}, whose behavior we study empirically in \cref{sec:experiments}.

\head{Capacity-scheduled memory}
To capture this formally, we augment the associative memory of \cref{def:associative-memory} with an \emph{activation operator} $\G_t$ that exposes only a subset of the memory's capacity at step $t$. Writing the online update of \cref{eq:generic-update} with this operator, incremental activation restricts both the write and the read to the active subset while holding the inactive components fixed.
\begin{dfn}[Capacity-Scheduled Associative Memory]
\label{def:incremental-associative-memory}
Let $\G_t$ be an activation operator that selects a subset of the memory parameters at step $t$, with $g_t \in \{0,1\}^{\dim(\M)}$ the indicator of the selected components, and write $\M_{t-1}^{(g)} := \G_t(\M_{t-1})$ for the active subset. A capacity-scheduled associative memory updates online as
\begin{align}
\label{eq:pos_dependent_am}
\M_t = \M_{t-1} - \theta_t\, \G_t\!\bigl(\nabla\tilde{\mathcal{L}}(\M_{t-1}^{(g)};\, k_t, v_t)\bigr),
\qquad \text{so that} \quad
(\mathbf{1} - g_t)\odot \M_t = (\mathbf{1} - g_t)\odot \M_{t-1},
\end{align}
and reads from the active subset, $y_t = \mathrm{Read}(\M_{t-1}^{(g)}, q_t)$.
\end{dfn}
The constraint states the defining property of the paradigm: components outside the active set at step $t$ are \emph{locked}, neither written nor read, and retain exactly the values they held before. Setting $\G_t$ to the identity for all $t$ recovers the standard online memory of \cref{sec:preliminaries}. Importantly, this is a single online trajectory under a \emph{growing} activation set, not a fresh optimization at each step: early tokens are compressed into the components that were active when they arrived, and that compressed state persists as further components are unlocked.

This formulation supports the two goals directly. For most of the context the active set is a strict subset of the memory, which limits effective capacity and forces compression through an explicit bottleneck. As additional components are unlocked, they supply fresh degrees of freedom that absorb later inputs, reducing interference with what earlier tokens have already written.

\head{What ``capacity'' means, and what it does not}
It helps to fix a concrete picture of the active subset before the specific schedule of \cref{sec:proteus}. For a matrix-valued memory $\M \in \mathbb{R}^{d_k \times d_v}$, the natural unit of capacity is a contiguous block of coordinates (e.g., a group of rows or columns), and activating a subset means exposing only some of these blocks to reads and writes while the rest stay locked. Two distinctions are worth drawing. First, the total memory is \emph{fixed}: we schedule which parts of a constant-size state are active, rather than growing the state itself, which sets our paradigm apart from approaches whose memory expands with sequence length (\cref{sec:related_work}). Second, activation is monotone over time: locked components are \emph{unlocked} as the context grows, rather than being permanently removed as in pruning, so no capacity is discarded and all of it becomes available by the end.

\head{Generality}
Although we have stated \cref{def:incremental-associative-memory} for the recurrent memory state, the activation operator $\G_t$ acts on nothing more specific than a set of capacity-bearing parameters. The same principle therefore applies wherever capacity is compressed into parameters over a context flow, including, through the Nested Learning view of \cref{sec:nl-pretraining}, the parameters of the model itself. This lets a single paradigm cover both the online memory (\cref{sec:proteus_memory}) and MLP blocks (\cref{sec:proteus_pretraining}). Incremental activation can be instantiated in many ways, differing in how $\G_t$ maps position to an active subset; in this work we study a simple, deterministic schedule and leave richer, data-dependent activation policies to future work.

\section{\model{}}
\label{sec:proteus}
We now introduce \model{}, our concrete realization of incremental memory activation (\cref{def:incremental-associative-memory}). We first present it for \emph{memory-based} recurrent sequence models, where an explicit associative memory state is updated online over the sequence. We then note that, through the Nested Learning view of \cref{sec:nl-pretraining}, the same principle applies to a model's parameters, letting \model{} extend to architectures such as Hope-Attention.

\subsection{\model{} for Memory}
\label{sec:proteus_memory}
\model{} instantiates the capacity-scheduled memory of \cref{def:incremental-associative-memory} with a simple block-wise activation operator. It requires two ingredients: a \textbf{partition} of the memory into blocks, and a deterministic \textbf{schedule} that activates them over the context. We describe each, then give the resulting gated update and read.

\head{Block partition and gating operator}
We partition the memory parameters into $E$ equal-sized contiguous blocks. Recall from \cref{def:incremental-associative-memory} that the activation operator $\G_t$ is realized by an elementwise mask $g_t \in \{0,1\}^{\dim(\M)}$; for \model{}, this mask enables exactly the parameters belonging to the currently active blocks, so that
\begin{align}\label{eq:gated_state}
    \G_t(\M) \;\coloneqq\; g_t \odot \M,
    \qquad\text{and in particular}\qquad
    \M_{t-1}^{(g)} \;=\; \G_t(\M_{t-1}) \;=\; g_t \odot \M_{t-1},
\end{align}
where $\odot$ is elementwise multiplication. Inactive blocks are masked out (locked) and participate in neither retrieval nor learning. Because most modern neural memories parameterize $\M$ as a linear map or an MLP, this multiplicative gating is all that is needed to control effective capacity, at no additional cost.

\head{Gated update and read}
Recall the generic online update and read of \cref{eq:generic-update}. \model{} instantiates the capacity-scheduled update of \cref{def:incremental-associative-memory} by applying the gradient only to active components, leaving locked blocks unchanged:
\begin{align}\label{eq:proteus_update}
\M_t
= \M_{t-1}
-\theta_t \bigl(g_t \odot \nabla \tilde{\mathcal{L}}(\M_{t-1}^{(g)}; k_t, v_t)\bigr),
\qquad \M_{t-1}^{(g)} = \G_t(\M_{t-1}).
\end{align}
Equivalently, \model{} updates the gated state and embeds it back into the full memory with locked components preserved,
\begin{align}\label{eq:proteus_update_equiv}
\M_t
= \M_{t-1} + \G_t\!\bigl(\delta \M_t\bigr),
\qquad \delta \M_t \coloneqq -\theta_t\, \nabla \tilde{\mathcal{L}}(\M_{t-1}^{(g)}; k_t, v_t).
\end{align}
Retrieval is gated the same way, reading only from the active subspace:
\begin{align}\label{eq:gated_read}
    y_t \;=\; \mathrm{Read} \bigl(\M_{t-1}^{(g)}, q_t\bigr)
    \;=\; \mathrm{Read} \bigl(\G_t(\M_{t-1}), q_t\bigr).
\end{align}
Both writes and reads are thus confined to the active blocks, while locked blocks remain untouched until activated. Although we write the update for gradient descent, the same gating applies unchanged to any update rule $\mathrm{Upd}(\cdot)$; only the active components are ever modified.

\head{Memory expansion schedule}
It remains to specify how position $t$ maps to a set of active blocks. We adopt a simple deterministic schedule that increases the number of active blocks uniformly across the context. Let $N$ be the maximum context length and $E$ the number of blocks, and assume $\dim(\M)$ is divisible by $E$ (otherwise the last block absorbs the remainder). With block size $d \coloneqq \dim(\M)/E$, define the step length
\begin{align}
    \Delta \;\coloneqq\; \max\Bigl(1,\;\bigl\lfloor \tfrac{N}{E} \bigr\rfloor \Bigr).
\end{align}
At step $t \in \{1,\dots,N\}$ we activate the first
$k(t) \coloneqq \min\!\left(E,\; 1 + \left\lfloor \frac{t-1}{\Delta} \right\rfloor \right)$
blocks, so the active fraction grows in $E$ uniform increments. The mask $g_t$ is the corresponding prefix mask:
\begin{align}
    g_t[j] \;=\;
    \begin{cases}
        1, & 1 \le j \le k(t)\, d,\\
        0, & \text{otherwise.}
    \end{cases}
\end{align}
Thus tokens in $[1,\Delta]$ read and write only the first $1/E$ of the memory, tokens in $[\Delta\!+\!1,2\Delta]$ the first $2/E$, and so on, until the full memory is available after the final expansion step (\cref{fig:proteus-arc}). We set $N$ to the training context length ($8$K) and keep this schedule fixed at evaluation; beyond $N$ the full memory is active, so \model{} adds no gating at inference lengths past the training window.

\subsection{Extension to Hope and MLP Blocks}
\label{sec:proteus_pretraining}
The same scheduling principle is not specific to the recurrent state. As set up in \cref{sec:nl-pretraining}, the Nested Learning view~\citep{behrouz2025nested} casts MLP blocks as an instances of associative memory with arbitrary objective (e.g., next token-prediction in language modeling) that compresses data into a MLP's parameters. Following the parallel drawn there, an over-parameterized MLP early in training has ample capacity to fit its first samples without compressing them; scheduling which parameters are active over the course of training applies the same early-bottleneck, fresh-capacity idea to the parameters themselves. We treat this as a lightweight extension rather than a separate contribution, and use it to reach architectures beyond recurrent memory.

\begin{table*}[t]
\centering
\caption{\textbf{Language modeling and commonsense reasoning results.}
We report perplexity on Wikitext and LAMBADA, and zero-shot accuracy on LAMBADA and seven commonsense benchmarks; Avg.\ denotes the mean accuracy across these eight accuracy columns.
Results are shown for 760M (50B tokens) and 1.3B (100B tokens) models, comparing each backbone with and without \model{} (shaded rows). Best average per group in \textbf{bold}.}
\label{tab:lm_results}
\centering
\resizebox{1.0\linewidth}{!}{
\centering
\begin{tabular}{l|c c|c c c c c c c c c}
\toprule
\textbf{Model}  & \textbf{Wiki.}  &  \textbf{LMB.} &  \textbf{LMB.} & \textbf{PIQA} &    \textbf{Hella.} & \textbf{Wino.} & \textbf{ARC-e} &  \textbf{ARC-c} &  \textbf{SIQA}  & \textbf{BoolQ} &  \textbf{Avg.} \\
 & ppl $\downarrow$  &  ppl $\downarrow$  &  acc $\uparrow$  & acc $\uparrow$ &   acc\_n $\uparrow$  & acc $\uparrow$  & acc $\uparrow$ & acc\_n $\uparrow$ &  acc $\uparrow$  & acc $\uparrow$ &   $\uparrow$  \\
\midrule
\midrule
\multicolumn{12}{c}{760M params / 50B tokens} \\
\midrule
 Transformer++ & 22.08 & 22.41 & 38.2 & 68.9 & 44.1 & 56.8 & 67.0 & 34.9 & 40.2 & 62.4 & 51.56\\
 Hope-Attention & 20.67 & 21.36 & 39.5 & 70.4 & 50.1 & 56.3 & 66.9 & 37.5 & 40.7 & 63.8 & 53.15\\
 \rowcolor{mygray} \qquad +\model & 19.87 & 19.72 & 40.5 & 70.2 & 51.9 & 58.2 & 67.3 & 38.5 & 41.4 & 63.9 & \textbf{53.99}\\
 \midrule
 RetNet & 23.54 & 23.87 & 35.7 & 66.4 & 42.8 & 53.6 & 64.8 & 33.1 & 38.7 & 57.4 & 49.06 \\
 SWLA ($c=2$) & 22.76 & 22.85 & 36.7 & 67.5 & 44.2 & 54.7 & 64.6 & 34.3 & 39.9 & 59.1 & 50.12\\
 \rowcolor{mygray} \qquad +\model & 21.82 & 20.88 & 37.5 & 67.1 & 44.9 & 54.7 & 66.0 & 35.8 & 41.0 & 60.2 & \textbf{50.90}\\ 
 \midrule
 DeltaNet & 22.89 & 23.13 & 37.2 & 68.0 & 45.2 & 52.9 & 65.4 & 33.4 & 40.2 & 59.8 & 50.26\\
 Comba & 21.94 & 21.77 & 38.4 & 67.1 & 47.1 & 53.0 & 65.9 & 35.5 & 40.8 & 63.7 & 51.43\\
 \rowcolor{mygray} \qquad +\model & 21.32 & 21.20 & 39.1 & 67.9 & 47.6 & 54.7 & 66.2 & 35.4 & 41.6 & 64.7 & \textbf{52.15}\\
 \midrule
 Titans & 20.92 & 21.28 & 39.3 & 69.2 & 49.9 & 53.1 & 67.3 & 36.9 & 41.7 & 63.8 & 52.65\\
 \rowcolor{mygray} \qquad +\model & 20.44 & 20.93 & 40.0 & 69.5 & 50.7 & 56.1 & 67.2 & 36.8 & 42.6 & 64.0 & \textbf{53.36}\\
 \midrule
 \toprule
 \multicolumn{12}{c}{1.3B params / 100B tokens} \\
 \midrule
 Transformer++ & 17.04 & 17.42 & 45.3 & 73.1 & 51.8 & 59.3 & 71.0 & 38.2 & 43.5 & 64.3 & 55.81\\
 Hope-Attention & 15.66 & 13.39 & 48.2 & 72.9 & 55.0 & 59.8 & 72.2 & 38.7 & 43.3 & 65.4 & 56.93\\
 \rowcolor{mygray} \qquad +\model & 15.70 & 13.01 & 48.5 & 73.1 & 55.4 & 60.8 & 73.1 & 38.5 & 44.2 & 64.8 & \textbf{57.30}\\
 \midrule
 RetNet & 18.84 & 17.29 & 40.8 & 70.6 & 48.9 & 56.0 & 67.9 & 35.3 & 41.6 & 62.7 & 52.98 \\
 SWLA ($c=2$) &18.11 & 16.95 & 40.7 & 71.4 & 49.2 & 57.3 & 68.8 & 37.0 & 42.4 & 62.6 & 53.67\\
 \rowcolor{mygray} \qquad +\model & 18.07 & 17.18 & 40.6 & 71.9 & 50.5 & 57.8 & 69.4 & 37.1 & 43.2 & 63.1 & \textbf{54.20}\\
 \midrule
 DeltaNet & 17.39 & 17.02 & 40.0 & 71.4 & 50.1 & 53.9 & 68.3 & 36.6 & 43.3 & 60.7 & 53.04 \\
 Comba & 16.98 & 14.17 & 43.8 & 73.2 & 53.5 & 60.1 & 70.9 & 39.2 & 44.2 & 58.5 & 55.42 \\
 \rowcolor{mygray} \qquad +\model & 16.95 & 14.01 & 44.6 & 73.1 & 54.2 & 60.4 & 71.4 & 39.5 & 44.6 & 60.2 & \textbf{56.00}\\
\midrule
Titans & 15.36 & 13.18 & 50.9 & 74.0 & 54.6 & 57.3 & 72.2 & 40.9 & 42.6 & 63.1 & 56.95\\
\rowcolor{mygray} \qquad +\model & 14.94 & 13.03 & 51.7 & 75.9 & 54.6 & 60.1 & 72.3 & 41.7 & 43.9 & 63.8 & \textbf{58.00}\\
\bottomrule
\end{tabular}
}
\end{table*}

\head{Gated parameter updates}
Recall the generic parametric update $\bm{\theta}_{i+1} = \bm{\theta}_i - \eta_{i+1} \bm{e}_i$ from \cref{eq:nl-backprop}, where $\bm{e}_i$ is the error (surprise) supplied by the optimizer for the $i$-th sample. \model{} gates this update with a schedule mask $g_{i+1}$ over the MLP-block parameters $\{\boldsymbol{\theta}_i\}$:
\begin{align}\label{eq:proteus-pretrain}
    \boldsymbol{\theta}_{i+1} = \boldsymbol{\theta}_i - \eta_{i+1}\,\bigl( g_{i+1} \odot \boldsymbol{e}_i \bigr),
\end{align}
so that only unmasked parameters receive a gradient update for each sample. For a concrete optimizer such as AdamW, this becomes
\begin{align}\label{eq:proteus-adamw}
     \boldsymbol{\theta}_{i+1} = \alpha_{i+1}\,\boldsymbol{\theta}_i - \eta_{i+1}\, \frac{\hat{\boldsymbol{m}}_{i+1}}{\sqrt{\boldsymbol{\hat{v}}_{i+1}} + \varepsilon},
\end{align}
with $\hat{\boldsymbol{m}}_{i+1} = \boldsymbol{m}_{i+1}/(1-\beta_1^{\,i+1})$, $\hat{\boldsymbol{v}}_{i+1} = \boldsymbol{v}_{i+1}/(1-\beta_2^{\,i+1})$, decoupled weight decay $\alpha_{i+1}$, and gated moment updates
\begin{align} \nonumber
    \boldsymbol{m}_{i+1} &= \beta_1\, \boldsymbol{m}_{i} + (1 - \beta_1)\bigl( g_{i+1} \odot \nabla_{\boldsymbol{\theta}_i} L(\boldsymbol{\theta}_i, \boldsymbol{x}_{i+1}) \bigr), \\ \nonumber
    \boldsymbol{v}_{i+1} &= \beta_2\, \boldsymbol{v}_{i} + (1 - \beta_2)\bigl( g_{i+1} \odot \nabla_{\boldsymbol{\theta}_i} L(\boldsymbol{\theta}_i, \boldsymbol{x}_{i+1}) \bigr)^{2}.
\end{align}
Only the parameters selected by $g_{i+1}$ are updated; scheduling $\{g_i\}$ over training then supplies fresh capacity for later data while avoiding full over-parameterization early on.

\head{Hope-Attention}
As a proof of concept, we apply \model{} to the Hope architecture~\citep{behrouz2025nested}, whose backbone resembles a Transformer but replaces the single post-attention MLP with a chain of MLP blocks updated at different frequencies. Given per-block update frequencies $f_i$ and input $\vx$, its output is
\begin{align}\label{eq:output-cms}
    &\vk_t = \vx_t W_{\vk}, \qquad \vv_t = \vx_t W_{\vv}, \qquad \vq_t = \vx_t W_{\vq},\\
    &\boldsymbol{h}_t = \texttt{Attn}\bigl(\{\vk_i\}_{i=1}^{t}, \{\vv_i\}_{i=1}^{t}, \vq_t \bigr), \\
    &\vy_t = \texttt{MLP}^{(f_k)}\bigl(\texttt{MLP}^{(f_{k-1})}(\cdots \texttt{MLP}^{(f_1)}(\boldsymbol{h}_t))\bigr),
\end{align}
where the $\ell$-th block's parameters are updated every $C^{(\ell)}$ steps. Applying \model{} here amounts to gating these MLP-block updates with the schedule mask of \cref{eq:proteus-pretrain}, so that each block's parameters are activated progressively over training. For language modeling, as a common standard, we use Next-Token Prediction with AdamW throughout.

\section{Experiments}
\label{sec:experiments}

\head{Experimental Setup}
We apply our framework to four families of high-performing models: Hope-Attention~\citep{behrouz2025nested}, Sliding-Window Linear Attention (SWLA)~\citep{behrouz2025atlas}, Comba~\citep{hu2025comba0}, and Titans~\citep{behrouz2024titans}. Unless otherwise noted, we set the number of partition blocks in \model{} to $E=16$; we ablate this choice in \cref{sec:ablations}. For reference, we also report results for Transformer++~\citep{touvron2023llama}, RetNet~\citep{sun2023retentive}, and DeltaNet~\citep{schlag2021linear}. All models are trained on FineWeb~\citep{penedo2024fineweb} with an 8K training context window. We consider two parameter scales (760M and 1.3B), training on 50B and 100B tokens respectively, and report perplexity on held-out Wikitext and LAMBADA. We use AdamW with a learning rate of $4 \times 10^{-4}$ and a cosine annealing schedule, a batch size of $0.5$M tokens, and weight decay of $0.1$. Downstream, we evaluate zero-shot on Wikitext~\citep{merity2017pointer}, LAMBADA (LMB)~\citep{paperno-etal-2016-lambada}, PIQA~\citep{bisk2020piqa}, HellaSwag~\citep{zellers2019hellaswag}, WinoGrande~\citep{sakaguchi2021winogrande}, ARC-Easy (ARC-e) and ARC-Challenge (ARC-c)~\citep{clark2018think}, SIQA~\citep{sap-etal-2019-social}, and BoolQ~\citep{clark-etal-2019-boolq}.

\begin{figure*}[t]
    \centering
    \includegraphics[width=\linewidth]{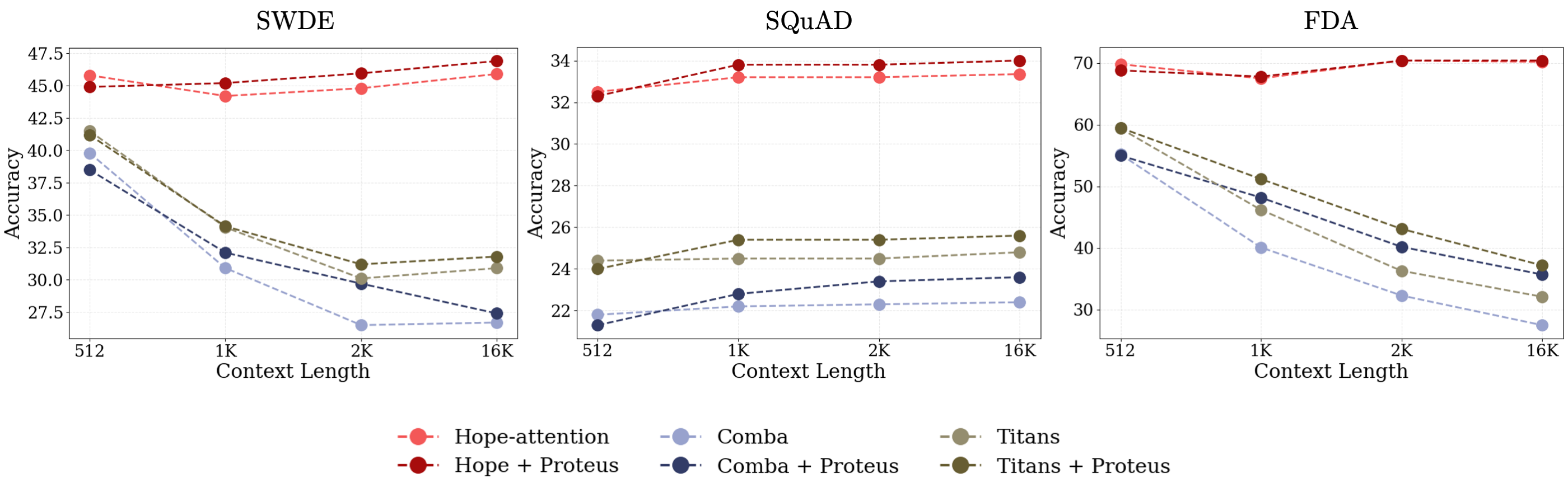}
    \caption{\textbf{Long-context retrieval.} Accuracy vs.\ context length on SWDE~\citep{lockard2019openceres}, SQuAD~\citep{rajpurkar2016squad}, and FDA~\citep{arora2024simple} for Hope-Attention, Comba, and Titans. \model{} consistently improves robustness at longer contexts, mitigating degradation (notably for Comba and Titans) and improving Hope-Attention across lengths.}
    \label{fig:longbench-plot}
\end{figure*}

\subsection{Language Modeling and Common-Sense Reasoning}
\label{exp:language_modeling}
Results for \model{} variants of the four backbones, together with additional baselines, are summarized in \autoref{tab:lm_results} at two scales. Across all four families and both scales, adding \model{} consistently improves average downstream accuracy over the corresponding base model, and lowers perplexity in nearly all settings.

At 760M, \model{} improves the average commonsense score for every backbone (Hope-Attention: $53.15 \rightarrow 53.99$; SWLA: $50.12 \rightarrow 50.90$; Comba: $51.43 \rightarrow 52.15$; Titans: $52.65 \rightarrow 53.36$), while also reducing perplexity in each case. The best overall 760M result is achieved by Hope-Attention+\model{}, with the lowest perplexities (Wiki.\ $19.87$, LMB.\ $19.72$) and the highest average accuracy ($53.99$). At 1.3B, the gains persist and are most pronounced for the strongest memory model: Titans+\model{} attains the best overall performance in the table, with the lowest perplexities (Wiki.\ $14.94$, LMB.\ $13.03$) and the highest average accuracy ($58.00$). These improvements hold even against strong attention-based baselines such as Transformer++, indicating that the benefit is not tied to a single architecture but to how memory is managed over the context.

The gains are consistent with \model{}'s two intended effects (\cref{sec:proteus}): the early capacity bottleneck yields cleaner compression, which helps even at the 8K training context, while the fresh capacity unlocked later aids retention as context grows. We examine the latter directly in the long-context experiments below.

\begin{table*}
\centering
\caption{
    \textbf{Needle-In-A-Haystack (NIAH) results.} Single-needle tasks at three levels of difficulty---S-NIAH-1 (pass-key retrieval), S-NIAH-2 (numerical needle), and S-NIAH-3 (UUID-based needle)---together with multi-key (MK), multi-query (MQ), and multi-value (MV) settings, at context lengths 4K, 8K, and 16K. Each backbone is shown with and without \model{} (shaded rows).}
    \label{tab:niah}
    \resizebox{0.75\linewidth}{!}{
    \centering
    \begin{tabular}{lccccccccc}
    \toprule
     & \multicolumn{3}{c}{\textbf{S-NIAH-1}}  & \multicolumn{3}{c}{\textbf{S-NIAH-2}} & \multicolumn{3}{c}{\textbf{S-NIAH-3}}  \\
    & \multicolumn{3}{c}{(pass-key retrieval)} & \multicolumn{3}{c}{(number in haystack)} & \multicolumn{3}{c}{(uuid in haystack)}  \\
    \cmidrule(lr){2-4} \cmidrule(lr){5-7} \cmidrule(lr){8-10}
    \textbf{Model}  & 4K  & 8K & 16K & 4K & 8K & 16K & 4K & 8K & 16K  \\
    \midrule
    Transformer     & 97.8 & 86.2 & 83.8 & 100  & 99.2 & 96.4 & 82.2 & 72.0 & 48.4 \\
    Hope-Attention  & 100  & 100  & 100  & 100  & 99.4 & 96.8 & 83.4 & 72.8 & 50.0 \\
    \rowcolor{mygray} \qquad +\model & 100 & 100 & 100 & 100 & 99.4 & 97.2 & 83.2 & 73.2 & 54.4 \\
    \midrule
    Comba           & 100  & 100  & 99.4 & 92.6 & 47.2 & 13.4 & 62.4 & 13.8 & 7.4 \\
    \rowcolor{mygray} \qquad +\model & 100 & 100 & 99.2 & 91.8 & 49.4 & 21.2 & 60.8 & 18.2 & 10.8 \\
    \midrule
    Titans          & 100  & 100  & 100  & 99.6 & 84.6 & 69.4 & 74.2 & 42.8 & 21.4 \\
    \rowcolor{mygray} \qquad +\model & 100 & 100 & 100 & 99.2 & 85.2 & 74.2 & 74.2 & 44.0 & 29.8 \\
    \bottomrule
    \addlinespace
    \toprule
    & \multicolumn{3}{c}{\textbf{MK-NIAH-1}} & \multicolumn{3}{c}{\textbf{MQ-NIAH}}& \multicolumn{3}{c}{\textbf{MV-NIAH}} \\
    & \multicolumn{3}{c}{(multi-key line retrieval)} & \multicolumn{3}{c}{(multi-query)} & \multicolumn{3}{c}{(multi-value)}  \\
    \cmidrule(lr){2-4} \cmidrule(lr){5-7} \cmidrule(lr){8-10}
    \textbf{Model}  & 4K & 8K & 16K & 4K & 8K & 16K & 4K & 8K & 16K  \\
    \midrule
    Transformer     & 80.0 & 79.4 & 60.8 & 57.4 & 46.2 & 28.8 & 36.6 & 33.8 & 20.2 \\
    Hope-Attention  & 81.4 & 85.2 & 62.2 & 62.0 & 47.6 & 30.6 & 37.2 & 36.2 & 23.0 \\
    \rowcolor{mygray} \qquad +\model & 80.8 & 85.2 & 64.8 & 62.2 & 48.4 & 34.2 & 36.6 & 36.0 & 25.8 \\
    \midrule
    Comba           & 22.8 & 20.4 & 10.4 & 22.2 & 16.0 & 6.2  & 15.8 & 14.2 & 6.6 \\
    \rowcolor{mygray} \qquad +\model & 23.4 & 20.2 & 14.2 & 23.8 & 16.6 & 8.4 & 14.8 & 14.0 & 8.2 \\
    \midrule
    Titans          & 27.0 & 22.8 & 11.8 & 23.2 & 18.4 & 10.2 & 24.8 & 15.4 & 8.8 \\
    \rowcolor{mygray} \qquad +\model & 28.4 & 22.6 & 16.8 & 23.2 & 18.6 & 12.0 & 24.6 & 14.8 & 10.2 \\
    \bottomrule
    \end{tabular}
    }
\end{table*}

\subsection{Needle-in-Haystack}
\label{sec:needle_in_haystack}

We evaluate long-context retrieval on Needle-in-a-Haystack (NIAH); results are in~\cref{tab:niah}. The pattern matches our thesis: \model{}'s gains are concentrated at longer contexts and on harder variants, while at short contexts where the base model already saturates (e.g., S-NIAH-1, near-perfect for all models), \model{} is essentially neutral. Where the base model degrades, the improvements are substantial. At 16K, \model{} improves Titans on S-NIAH-3 ($21.4 \rightarrow 29.8$) and S-NIAH-2 ($69.4 \rightarrow 74.2$), and Comba on S-NIAH-2 ($13.4 \rightarrow 21.2$). On multi-needle benchmarks at 16K, it improves MK-NIAH-1 (Titans $11.8 \rightarrow 16.8$), MQ-NIAH (Hope-Attention $30.6 \rightarrow 34.2$), and MV-NIAH (Hope-Attention $23.0 \rightarrow 25.8$). Across the table, the largest gains appear precisely where the context is longest and the base model weakest. Note that the schedule completes within the 8K training window, so at 16K no further blocks are unlocked: the improvement there reflects the cleaner, better-compressed state built under the early bottleneck, which degrades more gracefully as the context extends beyond the training length.

\subsection{Long Context Understanding}
We now evaluate \model{} on long-context understanding, using two families of tasks: recall-intensive retrieval, and the LongBench benchmark.

\head{Retrieval Tasks}
Following recent studies~\citep{hu2025comba0, yang2024gated, arora2024simple, behrouz2025atlas}, we evaluate \model{} on the Hope-Attention, Comba, and Titans baselines on recall-intensive tasks from \citet{lockard2019openceres, rajpurkar2016squad, arora2024simple}; results are in \autoref{fig:longbench-plot}. Across all datasets and baselines, \model{} is more robust at longer context lengths and improves length extrapolation. The effect is clearest for Comba and Titans, whose accuracy degrades sharply with context under the base model and is markedly stabilized by \model{}; for the already-strong Hope-Attention, \model{} improves accuracy across all lengths. Crucially, the gain over the baseline grows with context length, matching the trend seen in the NIAH results.

\head{LongBench Benchmark}
Finally, we evaluate \model{} on LongBench~\citep{bai2023longbench0}; results are in \autoref{tab:LongBench-table}. \model{} improves the average score for all three backbones and improves the large majority of individual tasks, without adding parameters or memory. Together with the retrieval and NIAH results, this supports our central claim: treating memory capacity as static is suboptimal, and incremental activation is a simple, broadly applicable mechanism for long-context management.

\begin{table*}
    \centering
    \caption{\textbf{LongBench evaluation.} Performance on LongBench~\citep{bai2023longbench0} across six long-context tasks (Narrative, Qasper, MultiField, Hotpot, 2WikiMulti, Musique), reported for each backbone with and without \model{} (shaded rows); Average denotes the mean across the six tasks. \model{} improves the average score for all three backbones.}
    \label{tab:LongBench-table}
    \resizebox{0.90\linewidth}{!}{
        \centering
        \begin{tabular}{l|c c c c c c c}
        \toprule
            \textbf{Model} & \textbf{Narrative} & \textbf{Qasper} & \textbf{MultiField} & \textbf{Hotpot} & \textbf{2WikiMulti} & \textbf{Musique} & \textbf{Average} \\
            \midrule\midrule
            Hope-Attention & 12.1 & 9.4 & 19.2 & 20.6 & 26.7 & 6.3 & 15.72\\
            \rowcolor{mygray}  \qquad + \model & 12.6 & 10.2 & 20.9 & 21.5 & 27.3 & 7.4 & 16.65\\
            \midrule
            Comba & 7.6 & 10.4 & 15.9 & 14.6 & 23.1 & 6.7 & 13.05\\
            \rowcolor{mygray}  \qquad + \model & 7.8 & 10.8 & 16.2 & 14.5 & 23.2  & 6.9 & 13.23\\
            \midrule
            Titans & 8.0 & 10.3 & 17.4 & 15.6 & 24.8 & 6.7 & 13.8\\
            \rowcolor{mygray}  \qquad + \model & 8.4 & 10.4 & 17.9  & 16.4 & 25.1  & 6.7 & 14.15\\
        \bottomrule
        \end{tabular}
    }
\end{table*}

\section{Related Work}
\label{sec:related_work}

\head{Modern Linear Recurrent Neural Networks}
To address the quadratic complexity of Transformers, recent work has focused on efficient recurrent architectures that enable rapid training and inference~\citep{tiezzi2024resurgence, katharopoulos2020transformers, yang2024gatedattn, schlag2021linear}. Early linear recurrent models, including linear attention~\citep{katharopoulos2020transformers}, RetNet~\citep{sun2023retentive}, RWKV~\citep{peng2023rwkv}, and S5~\citep{smith2023simplified}, utilized data-independent transition matrices with Hebbian update mechanisms. These evolved into second-generation architectures incorporating input-dependent parameters~\citep{hasani2023liquid, peng2024eagle, gu2024mamba, yang2024gated, yang2024gatedattn, zhao2026fast} and more expressive memory rules based on the delta rule~\citep{schlag2021linear, yang2024gated, liu2024longhorn, peng2025rwkv7, hu2025comba0}. Recent iterations have extended these mechanisms to use deep memory modules using momentum-based update rules~\citep{behrouz2024titans}, with non-linearity~\citep{sun2024learning, tang2026modular}, or alternative objectives~\citep{behrouz2025nested, behrouz2025Miras, kuratov2026gradmem, wang2026beyond, tang2026modular}. Other works focus on more powerful learning updates, including OmegaNet~\citep{behrouz2025atlas} and Oja's rules~\citep{irie2022neural} that extend upon delta and Hebbian rules, respectively. Notably, \citet{siems2025deltaproduct} recently proposed applying multiple gradient descent steps per token to enhance state tracking. Parallel to these linear approaches, significant progress has been made in optimizing the training and performance of RNNs with non-linear recurrence~\citep{behrouz2025Miras, csordas2024recurrent, merrill2024the, lim2024parallelizing, schone2025implicit, li2026tnt, von2023uncovering, gonzalez2024towards, behrouz2025nested, kuratov2026gradmem, irie2022modern, wang2026beyond}. Recent studies such as log-linear attention and memory caching have focused on designing recurrent models whose effective memory state grows with sequence length~\citep{behrouz2026memory, guo2025log, wei2025rat}. Such approaches are fundamentally different from ours: \model{} keeps a fixed-size memory and activates it incrementally, whereas these models grow the memory itself, targeting a different aspect of recurrent computation. They are orthogonal to our contribution, and combining growing-memory techniques with \model{}---so that a model both incrementally activates its memory parameters and expands its memory once all are active---is an interesting future direction.

\head{Adaptive Computation}
A parallel line of work reduces cost by allocating computation non-uniformly rather than spending full capacity on every input. This idea has a long history, from conditional computation~\citep{bengio2015conditional, graves2016adaptive} to early-exiting methods that halt processing for easy inputs~\citep{elbayad2019depth, schuster2022confident, mofakhami2024performance}. More recently, Mixture-of-Depths~\citep{raposo2024mixtureofdepths} routes only a subset of tokens through each layer's full computation, and Mixture-of-Recursions (MoR)~\citep{bae2026mixture} extends this to recursive Transformers, using lightweight routers to assign token-specific recursion depths while caching key--value pairs only for active tokens. These methods are related to \model{} in spirit, since both avoid applying the model's full capacity uniformly; however, they allocate \emph{compute across tokens} in a Transformer, whereas \model{} schedules \emph{effective memory capacity across positions} in a recurrent state. A further distinction is that their routing is \emph{data-dependent} and learned, while our schedule is a fixed function of position. Notably, MoR finds that learned recursion depth tracks the contextual predictability of each token; this suggests that extending \model{} with a learned, data-dependent activation policy---adapting capacity to content rather than position alone---is a promising direction.

\head{Non-Uniform Capacity Allocation}
Our work is part of a broader move away from allocating model capacity uniformly. Classical work established that controlling effective capacity aids generalization~\citep{guyon1991structural, hansen2001model, ding2018model}, and recent evidence shows that high-capacity models trained by empirical risk minimization tend to \emph{memorize} atypical examples rather than compress them into generalizable structure~\citep{bayat2024pitfalls}, which is precisely the failure mode incremental activation restricts early capacity to avoid. Most closely related, Tapered Language Models (TLMs)~\citep{bayat2026tapered} show that allocating more parameter capacity to earlier layers and less to later ones---under a fixed budget---improves language modeling across architectures, resonating with our thesis that capacity should be allocated non-uniformly rather than spread evenly across the model. 

\head{Associative Memory and Fast Weight Programs}
We ground our architecture in the principles of associative memory, originally framed by \citet{hopfield1982neural} as energy minimization for storing key-value pairs. While classical Hopfield networks were limited by low storage capacity, modern continuous variants have addressed these bottlenecks through dense associative memories and exponential kernels~\citep{krotov2016dense, krotov2021hierarchical, li2024expressive, lucibello2024exponential}, and have gained renewed prominence through their theoretical equivalence to the attention mechanism in Transformers~\citep{ramsauer2021hopfield, hu2024provably}. The related paradigm of treating linear layers as dynamic, writable memory was formalized in fast weight programmers~\citep{schmidhuber1992learning, schmidhuber1993reducing}, in which a slow network generates the weights of a fast network that processes the sequence. Two learning rules dominate this setting, Hebbian learning~\citep{hebb2005organization} and the delta rule~\citep{prados1989neural}, both of which let the network act as a key--value associative memory and have been studied extensively as efficient alternatives to standard attention~\citep{munkhdalai2017neural, munkhdalai2019metalearned, irie2021going, schlag2021linear, behrouz2025atlas, hasani2023liquid, peng2025rwkv7}.

\section{Conclusion}
We introduced \emph{incremental memory activation}, a simple paradigm for long-context sequence modeling in which the \emph{effective} capacity is progressively expanded over the context rather than exposed statically from the start.
We instantiated this idea in \model{}, a lightweight, block-wise gating mechanism that restricts both reads and writes to an active subset of the memory and unlocks fresh capacity over time, and, through the Nested Learning view, applies the same principle to a model's parameters.
Across four architectures (SWLA, Comba, Titans, and Hope-Attention) on language modeling and commonsense reasoning, and on needle-in-a-haystack, retrieval, and LongBench for the memory backbones we evaluate at length, \model{} consistently improves long-context robustness and length extrapolation, with the largest gains appearing at the longest context lengths.
That a single, architecture-agnostic mechanism helps across four distinct memory designs suggests the benefit stems from how capacity is allocated over the context, not from any one architecture: treating capacity as static is suboptimal, and scheduling it is a practical and broadly applicable tool for long-context computation.

\head{Limitations}
Our study has several limitations. First, the activation schedule is fixed and hand-designed: we grow the active set uniformly with position, and while our ablations show that finer partitions help, we do not characterize what schedule is optimal or how it should depend on the data or the memory's update rule. Second, the extension to MLP blocks is a proof of concept demonstrated on a single architecture, rather than a fully validated result. Third, \model{}'s gains are concentrated at longer contexts and harder tasks; where the base model already saturates, the effect is essentially neutral, so the mechanism is most useful precisely when long-context capacity is the bottleneck. 

\head{Future Directions}
Several directions follow naturally. A first is to replace our fixed schedule with a \emph{learned}, data-dependent activation policy, and to characterize the optimal schedule, including whether uniform growth is the right choice and how it depends on data statistics and the underlying update rule. A second is to combine incremental activation with growing-memory architectures such as memory caching~\citep{behrouz2026memory}, so that a model both activates its fixed memory over the context and expands that memory once all blocks are active. This pairing is a particularly natural fit: incremental activation governs how a fixed budget is spent within the training window, while a growing memory supplies genuinely new capacity beyond it, so the two mechanisms cover complementary regimes and together could extend the schedule indefinitely with context. A third is to extend the schedule beyond the training window at inference time, stretching $\Delta$ so that capacity continues to unlock at context lengths longer than those seen in training; whether a state compressed under the training schedule remains well-calibrated when the unlock points move is an open question. Finally, the same principle extends beyond recurrent memory to other parameter subsets and to continual adaptation settings, where scheduling which parameters are active over a long stream of experience may help balance fast adaptation against retention of past knowledge.

\clearpage
\printbibliography

\clearpage
\appendix

\begin{figure}
    \begin{minipage}{0.45\linewidth}
    \centering
    \includegraphics[width=\linewidth]{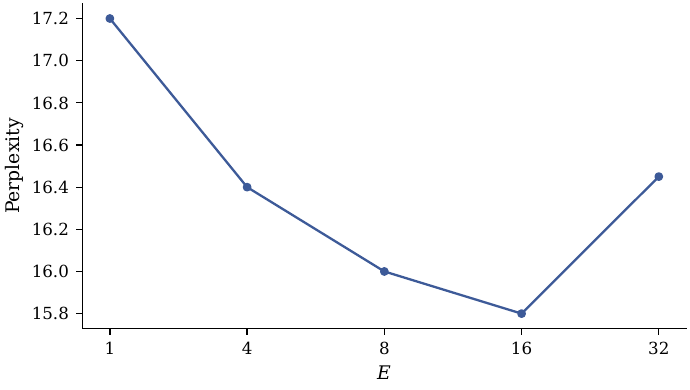}
    \caption{The effect of $E$ on the performance.}
    \label{fig:blocks}
    \end{minipage}
    \hfill
    \begin{minipage}{0.45\linewidth}
    \centering
    \includegraphics[width=\linewidth]{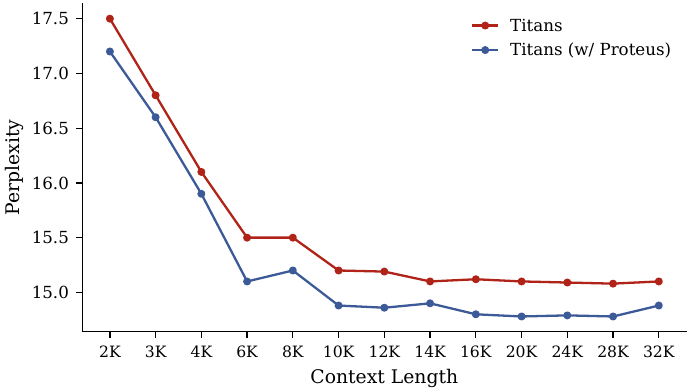}
    \caption{Perplexity-by-token-index comparison.}
    \label{fig:ppl}
    \end{minipage}
\end{figure}

\section{Ablations and Analysis}
\label{sec:ablations}

\head{Effect of the number of blocks}
The number of partition blocks $E$ is the central hyperparameter of \model{}, controlling how finely the memory's effective capacity is scheduled over the context. We ablate it by varying $E$ from $1$ to $32$, holding everything else fixed. Note that $E=1$ recovers the base model exactly: with a single block there is no scheduling, so the full memory is active from the first token. This makes the ablation a clean isolation of the scheduling effect, since only $E$ changes. \autoref{fig:blocks} reports validation perplexity as a function of $E$. Perplexity improves sharply from $E=1$ to $E=8$ and reaches its minimum at $E=16$, but degrades again at $E=32$. This non-monotonicity is precisely the tension anticipated in \cref{sec:incremental_memory_activation}: a finer partition sharpens the early bottleneck, which improves compression up to a point, but too fine a schedule leaves too little capacity active for the initial context, since at $E=32$ the first tokens read and write only $1/32$ of the memory, so the bottleneck begins to cost more than the compression it buys.

\head{Perplexity by token position}
To examine \emph{where} \model{}'s gains arise, we measure validation perplexity as a function of token position in the context, comparing a base model against its \model{} variant (\autoref{fig:ppl}). \model{} achieves lower perplexity at every position, indicating that the scheduled capacity helps throughout the context rather than trading early-context quality for late-context gains. The gap widens as the schedule progresses and is largest around the $8$K training context length, where the memory becomes fully active; beyond that point it narrows slowly but remains positive out to $32$K. This is consistent with the mechanism described in \cref{sec:proteus_memory}: the advantage accumulates while fresh capacity is still being unlocked, and the better-compressed state that results continues to pay off at lengths past the training window, where no further activation occurs.

\end{document}